\pdfoutput=1
\documentclass[11pt]{article}

\usepackage{naacl2021}

\usepackage[mode=buildnew]{standalone}
\usepackage[none]{hyphenat}
\usepackage{amsmath}
\usepackage{array}
\usepackage{arydshln}
\usepackage{booktabs}
\usepackage{empheq}
\usepackage{float}
\usepackage{graphicx}
\usepackage{hyperref}
\usepackage{latexsym}
\usepackage{makecell}
\usepackage{multirow}
\usepackage{pgfplots}
\usepackage{subcaption}
\usepackage{tabularx}
\usepackage{tikz}
\usepackage{url}
\usepackage{verbatim}

\usetikzlibrary{patterns, shapes.geometric, positioning, bayesnet}

\pgfplotsset{compat=newest}
\pgfplotsset{filter discard warning=false}
\newcolumntype{L}[1]{>{\raggedright\let\newline\\\arraybackslash\hspace{0pt}}m{#1}}
\newcolumntype{C}[1]{>{\centering\let\newline\\\arraybackslash\hspace{0pt}}m{#1}}
\newcolumntype{R}[1]{>{\raggedleft\let\newline\\\arraybackslash\hspace{0pt}}m{#1}}
\newcolumntype{Y}{>{\centering\arraybackslash}X}

\newcommand\xbf[1]{\bf #1}
\newcommand\xtt[1]{\tt #1}
\newcommand*\samethanks[1][\value{footnote}]{\footnotemark[#1]}

\title{Using Noisy Self-Reports to Predict Twitter User Demographics}

\author{
Zach Wood-Doughty\Thanks{ Equal contribution}, Paiheng Xu\samethanks,
Xiao Liu, Mark Dredze \\
Department of Computer Science \\
John Hopkins University, Baltimore, MD 21218 \\
{\tt zach@cs.jhu.edu,paiheng@jhu.edu,xliu119@jhu.edu,mdredze@cs.jhu.edu}
}

\begin{document}

\maketitle

\begin{abstract}
Computational social science studies often contextualize content analysis
within standard demographics.  Since demographics are unavailable on many
social media platforms (e.g.  Twitter), numerous studies have inferred
demographics automatically.  Despite many studies presenting proof-of-concept
inference of race and ethnicity, training of practical systems remains elusive
since there are few annotated datasets.  Existing datasets are small,
inaccurate, or fail to cover the four most common racial and ethnic groups in
the United States.  We present a method to identify self-reports of race and
ethnicity from Twitter profile descriptions.  Despite the noise of
automated supervision, our self-report datasets enable
improvements in classification performance on gold standard self-report survey
data.  The result is a reproducible method for creating large-scale training
resources for race and ethnicity.
\end{abstract}

\section{Introduction}

Contextualization of population studies with demographics forms a central analysis method within the social sciences. In domains such as political science or public health,
standard demographic panels in telephone surveys enable better analyses of opinions and trends.
Demographics such as age, gender, race, and location are often proxies for important socio-cultural groups. 
As the social sciences increasingly rely on computational analyses of online text data, 
the unavailability of demographic attributes hinders comparison of these studies to traditional methods~\cite{al2020linking,amir2019population,jiang2020not}.

Computational social science increasingly utilizes methods for the automatic inference of demographic attributes from social media, such as Twitter ~\cite{burger2011discriminating,chen2015comparative,ardehaly2017co,jung2018assessing,huang2019neural}.
Demographics factor into social media studies across domains such as
health, politics, and linguistics~\cite{o2010tweets,eisenstein2014diffusion}.
Off-the-shelf software packages support the inference of gender
and location~\cite{knowles2016demographer,dredze2013carmen,wang2019demographic}.

Unlike age or geolocation, race and ethnicity are sociocultural categories
with competing definitions and measurement
approaches \cite{comstock2004four,vargas2016documenting,culley2006transcending,andrus2021what}.
Despite this complexity, understanding race and ethnicity is crucial for
public health research~\cite{coldman1988classification,dressler2005race,fiscella2006use,elliott2008new,elliott2009using}.
Analyses that explore mental health on Twitter~\cite{loveys2018cross} should consider
racial disparities in healthcare~\cite{satcher2001mental,amir2019population}
or online interactions~\cite{delisle2019troll,burnap2016us}.
Despite the importance of race and ethnicity in these studies,
and multiple proof-of-concept classification studies,
there are no readily-available systems that can infer demographics
for the most common United States racial/ethnic groups.
This gap arises from major limitations for
all publicly-available data resources.

\begin{table*}[!t]
\centering
\small
\begin{tabular}{ c c c c  c c c c }
\toprule
{\bf Citation} & {\bf Annotation} & {\bf \% Missing} & {\bf \# Users} & {\bf \% W} & {\bf \% B} & {\bf \% H/L} & {\bf \% A} \\
\midrule
\citet{preoctiuc2015studying} & Survey       & 4.7  & 3572  & 80.8 &  9.5 &  6.1 &  3.6 \\
\citet{culotta2015predicting} & Crowdsourced & 60.0 &  308  & 50.0 & 19.5 & 30.5 &    0 \\
\citet{volkova2015predicting} & Crowdsourced & 36.5 & 3174  & 48.0 & 35.8 &  8.9 &  3.0 \\
\midrule
%                                                             White  Black   H/L   Asian
Total Matching Users            & Self-report & -   & 2.50M & 26.8 & 53.8  & 11.3 & 8.1  \\
Query-Bigram                    & Self-report & 8.1 & 112k  & 51.2 & 40.8  & 1.4  & 6.6  \\
Heuristic-Filter                & Self-report & 40.6& 135k  & 42.2 & 45.9  & 5.6  & 6.4  \\
Class-Balanced                  & Self-report & 0.0 & 31k   & 25.0 & 25.0  & 25.0 & 25.0 \\
\bottomrule
\end{tabular}
\caption{Previously-published Twitter datasets annotated for race/ethnicity and datasets collected in this work.
``\% Missing'' shows the percent of users that could not be scraped in 2019.
``\# Users'' shows the number users that are currently available.
The abbreviations W, B, H/L, and A corresponds to White, Black, Hispanic/Latinx, Asian respectively, which we use for the rest of the paper.
Per-group percentages are from non-missing data.
}
\label{tbl:existing_datasets}
\end{table*}

A high-quality dataset for this task has several desiderata.
First, it should cover enough categories to match standard demographics panels.
Second, the dataset must be sufficiently large to support training accurate systems.
Third, the dataset should be reproducible; Twitter datasets shrink as users delete or restrict accounts,
and models become less useful due to domain drift~\cite{huang2018examining}.

We present a method for automatically constructing a large Twitter dataset for race and ethnicity. 
Keyword-matching produces a large, high-recall corpus of Twitter users who potentially self-identify as a racial or ethnic group,
building on past work that considered self-reports~\cite{ardehaly2014using,beller2014belieber,coppersmith2014quantifying}.
We then learn a set of filters to improve precision by removing users who match keywords but
do not self-report their demographics.
Our approach can be automatically repeated in the future to update the dataset.
While our automatic supervision contains noise -- self-descriptions are
hard to identify and potentially unreliable -- our large dataset demonstrates
benefits when compared to or combined with previous crowdsourced
datasets. We validate this comparison on a gold-standard survey dataset of
self-reported labels~\cite{preoctiuc2018user}. We release our code
publicly\footnote{\tiny\url{https://bitbucket.org/mdredze/demographer}}. We
also release our collected datasets and trained models to researchers with
approval from an IRB or similar ethics board, contingent on compliance with our
data usage
agreement\footnote{\tiny\url{http://www.cs.jhu.edu/~mdredze/demographics-training-data/}}. 

\section{Ethical Considerations} \label{sec:ethics}

Complexities of racial identity raise ethical considerations, requiring discussion
of the benefits and harms of this work \cite{bentonetal2017ethical}.
The benefits are clear in settings such as public health;
many studies use social media data to research
health behaviors or support health-based interventions~\cite{paul2011you,sinnenberg2017twitter}.
These methods have transformed areas of public health
which otherwise lack accessible data~\cite{ayers2014could}.
Aligning social media analyses with traditional data sources requires
demographic information.

The concerns and potential harms of this work are more complex.
Ongoing discussions in the literature concern the need for informed consent from social media users~\cite{fiesler2018participant,marwick2011tweet,olteanu2019social}.
Twitter's privacy policy states that the company
``make[s] public data on Twitter available to the world,''
but many users may not be aware of the scope or nature of research conducted using their data 
\cite{mikal2016ethical}. Participant consent must be {\it informed},
and we should study users' comprehension of terms of service
when conducting sensitive research.
IRBs have applied established human subjects research regulations
in ruling that passive monitoring of social media data falls under public data exemptions.

While our data usage agreement prohibits such behavior, a malicious actor could
attempt to use predicted user demographics to track or harass minority
groups. Despite the severity of such a worst-case scenario, there are two
arguments why the benefits may outweigh the harms. 
First, if open-source methods and models were used for such malicious behavior,
platform moderators could simply incorporate those tools into combatting
any automated harassment.  Second, harassment against historically
disenfranchised groups is already extremely widespread. Open-source tools would
provide more good than harm in the hands of researchers or platform
moderators~\cite{jiang2020not}.  Recent work has show that women on Twitter, 
especially journalists and politicians, receive disproportionate amounts of
abuse~\cite{delisle2019troll}. On Facebook, advertisers have used the platform's
knowledge of users' racial identities to illegally discriminate when
posting job or housing ads~\cite{benner2019facebook,angwin2016facebook}.
To protect against misuse of our work, we follow 
Twitter's developer terms which prohibit efforts to
``target, segment, or profile individuals'' based on several sensitive categories,
including racial or ethnic origin, detailed in our data use agreement.
Predictions should not be analyzed to profile individual users but rather
must only be used for aggregated analyses.

Another concern of any predictive model for sensitive traits
is that a descriptive model could be interpreted as a prescriptive assessment~\cite{ho2015essentialism,crawford2017trouble}.
Individual language usage may also differ from population-level demographics patterns~\cite{bamman2014gender}.
Additionally, our datasets and models do not cover
smaller racial minorities (e.g. Pacific Islander) or the fine-grained
complexities of mixed-race identities. More fine-grained methods
are needed for many analyses, but current methods cannot support them.

Finally, we distinguish between biased models and biased applications.
Our models are imperfect; if we only analyze a small sample of users
and our models have high error rates, a difference that appears significant may
be an artifact of misclassifications. Any downstream application must
account for this uncertainty.

On the whole, we believe demographic tools provide significant benefits that justify the potential risks in their development.
We make our data available to other researchers, but with limitations.
We require that researchers 
comply with a data use agreement and
obtain approval by an IRB or similar ethics committee.
Our agreement restricts these tools to population-level analyses\footnote{Twitter's API ``restricted use cases'' explicitly permit aggregated analyses.}
and {\bf not} the analysis of individual users.
We exclude certain applications, such as targeting of individuals based on race or ethnicity.
Any future research that makes demographically-contextualized conclusions from
classifier predictions must explicitly consider ethical trade-offs specific to its application.
Finally, our analysis of social media for public health research has been IRB reviewed and deemed exempt (45 CFR 46.101(b)(4)).

\section{Datasets for Race and Ethnicity}
Our tools and analysis focus on the United States, where 
recognized racial categories
have varied over time \cite{hirschman2000meaning,lee2006rethinking}.
Current US census -- and many surveys -- record self-reported racial categories as
White, Black, American Indian, Asian, and Pacific Islander.
Surveys often frame ethnicity as Hispanic/Latinx origin or not;
however, there is not necessarily a clear distinction between
race and ethnicity~\cite{gonzalez2015hispanic,campbell2006categorical,cornell2006ethnicity}.
Individuals may identify as both a race and an ethnicity,
and 2\% of Americans identify as multi-racial~\cite{jones2001two}.
Because of the limited data availability,
we only consider the four largest race/ethnicity groups, which
we model as mutually exclusive: White, Black, Asian, and Hispanic/Latinx.
Our methodology could be extended to be more comprehensive,
but we do not yet have the means to validate more fine-grained or intersectional approaches. 

Table~\ref{tbl:existing_datasets}
lists three published datasets for race/ethnicity.
Since only user ids can be shared,
user account deletions over time cause substantial missing data.
Past work has taken varied approaches to annotate racial demographics.
\citet{culotta2015predicting} and \citet{volkova2015predicting} relied on manual annotation,
noting inter-annotator agreement estimated at 80\% and Cohen's $\kappa$ of 0.71, respectively.
Crowdsourced annotation assumes that racial identity can be accurately perceived by others,
an assumption that has serious flaws for gender and age~\cite{flekova2016analyzing,preoctiuc2017controlling}.
Rule-based or statistical systems for data collection can be
effective~\cite{burger2011discriminating,chang2010epluribus},
but raise concerns about selection bias: if we only label users who take a certain action,
a model trained on those users may not generalize to users who do not take that action~\cite{wood2017does}.

Gold-standard labels for sensitive traits requires individual survey responses,
but this yields small or skewed datasets due to the expense \cite{preoctiuc2018user}.
Our approach instead relies on automated supervision from racial self-identification
and minimal manual annotation to refine our dataset labels.
We are not the first to use users' self-identification to label Twitter users' demographics,
but past work has relied heavily either on restrictive regular expressions or
manual annotation~\cite{pennacchiotti2011machine,ardehaly2014using}.
Such work has also been limited to datasets of under 10,000 users.
We expand on previous work to construct a much larger dataset and evaluate it
via trained model performance on ground-truth survey data.

\begin{table}[!t]
\small
\centering
\setlength{\tabcolsep}{3pt}
\begin{tabular}{ c c c c c c c }
\toprule
                                  &  Raw   & Color & Plural & Bigram & Quote &  All           \\
\midrule
Precision                         &  76.7 & 78.6 & 76.7  & 82.5  & 78.6 & \textbf{86.8} \\ 
\midrule
\makecell{Removed\\ by filter} & -      & 314k  & 212k   & 281k   & 4k    & 784k           \\
\bottomrule
\end{tabular}
\caption{
Applying our {\tt HF} filters (\S~\ref{sec:data_collection}) individually and together.
Precision is on dev set from Appendix~\ref{sec:self-report}, after thresholding on self-report score.
}
\label{tbl:effect-tricks}
\end{table}

\section{Data Collection of Self-Reports} \label{sec:data_collection}
We construct a regular expression for terms associated with racial identity.
We select tweets from Twitter's 1\% sample from July 2011 to July 2019
in which the user's profile description contains one of the following racial keywords in English:
\texttt{black}, \texttt{african-american}, \texttt{white}, \texttt{caucasian}, \texttt{asian}, \texttt{hispanic}, \texttt{latin}, \texttt{latina}, \texttt{latino}, \texttt{latinx}.
While there are other terms that signify racial identity,
these match common survey panels~\cite{hirschman2000meaning}
and our empirical evaluation is limited because our survey dataset only covers four classes.
We omit self-reports that indicate a country of origin (e.g. ``Colombian'' or ``Chinese-American''),
smaller racial minorities (e.g. ``Native American'' or ``two or more races''),
or more ambiguous terms, leaving such groups for future work.
If a user appears multiple times, we use their latest description.

We select users whose profile descriptions contain a query keyword,
which heavily skews towards color terms (``white'', ``black'').
This produces 2.67M users, 2.50M of which match exactly one racial/ethnic class
(Table~\ref{tbl:existing_datasets}, ``Total Matching Users'').
While this is several orders of magnitude larger than existing datasets,
many user descriptions that match racial keywords are not racial self-reports.
We next consider approaches to filter these users' profile descriptions
to obtain three self-report datasets of different sizes and precisions.

For all three datasets, we use a model that assigns a ``self-report'' score
based on the likelihood that a profile contains a self-report.
We then use a binary cutoff to only include users with a high enough
self-report score. We obtain this score by leveraging lexical co-occurrence,
an important cue for word associations~\cite{spence1990lexical,church1990word}.
We combine relative frequencies of co-occurring words within a fixed window,
weighed by distance between query and co-occurring self-report words.
For example, if ``farmer'' is a self-report word, then
``Black farmer'' should score higher than ``Black beans farmer''
since the query and self-report word are closer.
We choose the window size and threshold for this score function on a
manually-labeled tuning set, after which our scoring function achieves 72.4\%
accuracy on a manually-labeled test set. Details on preprocessing and our
self-report score are in Appendices~\ref{sec:preprocessing} and
\ref{sec:self-report}.

Our first dataset  selects users with a bigram containing a racial keyword
followed by a ``person keyword.'' Our person keywords are:
\texttt{man}, \texttt{woman}, \texttt{person}, \texttt{individual},
\texttt{guy}, \texttt{gal}, \texttt{boy}, and \texttt{girl}
so this method matches users with descriptions containing bigrams such as ``Black woman'' or ``Asian guy.''
We expect this method to have high precision, but it has extreme label imbalance;
91\% of the users are labeled as either white or black.
From the Twitter 1\% sample, this dataset contains 122k users,
but only 112k users could be re-scraped in 2019.
We refer to this dataset as {\bf Query-Bigram (QB)}.

As {\bf QB} contains only 112k users,
we consider a less restrictive approach.
Our second dataset uses four heuristic filters to remove false positives from the original 2.67M users.
Many descriptions spuriously match ``black'' and ``white'' in addition to other colors,
so we filtered out all words from a color-list~\cite{berlin1991basic}.
Second, we filter out racial keywords followed by plural nouns
(e.g. ``white people''), using NLTK {\tt TweetTokenizer}~\cite{bird2009natural}
to obtain part-of-speech tags.
We curate a list of 286 Google bigrams that frequently contain a query but
are unlikely to be self-reports (e.g. ``black sheep,'')~\cite{michel2011quantitative}.
Finally, we ignore query words that appear inside quotation marks.
Table \ref{tbl:effect-tricks} shows how precision and dataset size change
as we apply these filters.
Applying all four gives a total of 1.72M users; after thresholding on
self-report score we are left with 228k users.
135k such users could be scraped in 2019, producing our
{\bf Heuristic-Filtered (HF)} dataset.

As {\bf QB} and {\bf HF} are quite imbalanced, we design a third dataset to
equally represent all four classes.
Across both our {\bf QB} and {\bf HF} datasets we have only 7,756 Hispanic/Latinx users
that we could scrape in 2019, making it our smallest demographic class.
We thus use our self-report scores to select the highest-scoring 7,756 users
from each of other classes, producing our {\bf Class-Balanced (CB)} dataset of 31k users.

\begin{table*}[!ht]
\newlength{\gap}
\setlength{\gap}{1em}
\small
\centering
\setlength{\tabcolsep}{4pt}
\begin{tabularx}{\linewidth}{C{8em} @{\hspace{\gap}} *{6}Y @{\hspace{\gap}} *{6}Y}
& \multicolumn{6}{@{}c@{}}{Imbalanced prediction}
& \multicolumn{6}{c}{Balanced prediction} \\
\addlinespace
& \multicolumn{2}{c}{Names} & \multicolumn{2}{c}{Unigrams} & \multicolumn{2}{@{}c@{}}{BERT}
& \multicolumn{2}{c}{Names} & \multicolumn{2}{c}{Unigrams} & \multicolumn{2}{c}{BERT} \\
\cmidrule(lr){2-3} \cmidrule(lr){4-5} \cmidrule(lr{\gap}){6-7} \cmidrule(l){8-9} \cmidrule(lr){10-11} \cmidrule(lr){12-13}
Dataset/Baseline & F1        & Acc\%       & F1       & Acc\% & F1       & Acc\% &
F1        & Acc\%       & F1       & Acc\% & F1       & Acc\% \\
\cmidrule(lr{\gap}){1-1} \cmidrule(r{\gap}){2-7} \cmidrule(r){8-13}
%                 name-f1   name-acc     unig-f1   unig-acc     bert-f1  bert-acc   ||  name-f1   name-acc      unig-f1  unig-acc     bert-f1     bert-acc      
Random         & .250      & 25.0      & .250     & 25.0      &.250      &25.0         &.250      &25.0       &.250      &25.0      & .250       & 25.0       \\
Majority       & .224      &\xbf{80.8} & .224     & 80.8      &.224      &\xbf{80.8}   &.100      &25.0       &.100      &25.0      & .100       & 25.0       \\
\addlinespace                                                                                                                                                   
Crowd          & .268 	   & 74.9      &.432      & \xbf{83.2}&\xbf{.402}& 74.8        &.213 	   &.322 	      &.343      &40.9      & .402       & 43.7       \\
\addlinespace                                                                                                                                                   
\xtt{QB}       &\xbf{.335} & 71.7      &.394      & 71.4      &.371      & 61.0        &.316 	   &.377 	      &.406      &46.5      & .461       & 48.3       \\
Crowd+\xtt{QB} & .331 	   & 74.3      &\xbf{.460}& 78.4      &.383      & 62.4        &.276 	   &.344 	      &.453      &47.6      & .484       & 50.1       \\
\addlinespace                                                                                                                                                    
\xtt{HF}       & .324 	   & 64.4      &.401      & 72.4      &.346      & 62.3        &.308 	   &.377 	      &.418      &47.3      & .408       & 44.1       \\ 
Crowd+\xtt{HF} & .198 	   & 54.0      &.449      & 76.9      &.360      & 62.1        &.149 	   &.233 	      &\xbf{.466}&\xbf{50.9}& .441       & 47.4       \\ 
\addlinespace                                                                                                                                                    
\xtt{CB}       & .299      & 49.4      &.300      & 43.3      &.285      & 39.0        &.379 	   &.381 	      &.463      &48.9      & .474       & 49.0       \\ 
Crowd+\xtt{CB} & .249 	   & 35.9      &.449      & 74.6      &.349      & 52.0        &\xbf{.386}&\xbf{.390} &.465      &48.9     & \xbf{.514} & \xbf{52.6}  \\ 
\cmidrule[.08em](lr{\gap}){1-1} \cmidrule[.08em](r{\gap}){2-7} \cmidrule[.08em](r){8-13}
\end{tabularx}
\caption{Experimental results for baseline methods, models trained on the crowdsourced datasets,
and models trained on our self-report datasets. The best result in each column is in bold.
Dataset abbreviations are defined in \S~\ref{sec:data_collection}.
``+'' indicates a combined dataset of crowdsourced data plus our self-report data.
Section~\ref{sec:experiments} and Appendix~\ref{app:hyperparameter} contain the training and evaluation details.
}
\label{tbl:results}
\end{table*}

\begin{table}[!h]
\small
\centering
\begin{tabular}{ c  c c c c}
\toprule
& \multicolumn{4}{c  }{Imbalanced} \\
\makecell{Method} & W    & B     & H/L    & A   \\ 
\midrule                                        
Random             & 25.0 & 25.0  & 25.0 & 25.0 \\ 
Majority           & 100. &  -    &  -   &  -   \\ 
\midrule                                       
Crowd              & 95.1 & 49.8  & 0.9  & 19.1 \\ 
\midrule                                      
{\tt QB}           & 77.7 & 74.0  & 5.4  & 30.1 \\ 
Crowd+{\tt QB}     & 86.5 & 66.5  & 13.7 & 29.2 \\ 
\midrule                                     
{\tt HF}           & 78.9 & 74.3  &  7.4 & 25.6 \\ 
Crowd+{\tt HF}     & 84.2 & 72.1  & 14.7 & 24.8 \\ 
\midrule                                    
{\tt CB}           & 41.1 & 77.1  & 16.7 & 51.3 \\ 
Crowd+{\tt CB}     & 81.1 & 68.7  & 20.1 & 30.1 \\ 
\midrule
 & \multicolumn{4}{c }{Balanced} \\
\makecell{Method} &    W    & B     & H/L   & A     \\
\midrule                                                
Random             & 25.0 & 25.0 & 25.0 & 25.0  \\     
Majority           & 100. &  -   &  -   &  -   \\
\midrule                                               
Crowd              & 95.6 & 51.3 & 15.0 & 1.8  \\
\midrule                                               
{\tt QB}           & 75.2 & 75.2 & 5.3  & 30.1 \\
Crowd+{\tt QB}     & 76.1 & 67.3 & 25.6 & 21.2 \\
\midrule                                               
{\tt HF}           & 77.9 & 77.0 & 8.9 & 25.6 \\
Crowd+{\tt HF}     & 87.6 & 73.5 & 15.9 & 26.5 \\
\midrule                                               
{\tt CB}           & 41.6 & 82.3 & 20.4 & 51.3  \\
Crowd+{\tt CB}     & 72.6 & 72.6 & 19.5 & 31.0 \\
\bottomrule
\end{tabular}
\caption{Class-specific accuracy for Unigram models. Dashes indicate 0\% accuracy.
In general, the more class-imbalanced a dataset is, the worse it does on the smaller
classes. In the imbalanced setting, the Unigram model trained on the Crowd
dataset achieves the best accuracy solely due to its 95.1\% accuracy
on the users labeled as White.
}
\label{tbl:class_specific_accuracy}
\end{table}

\section{Experimental Evaluation} \label{sec:experiments}

We now conduct an empirical evaluation of our noisy self-report datasets.
Showing that our datasets produce accurate classifiers demonstrates
the value of our noisy self-report method for dataset construction.
We train supervised classifiers on both our and existing datasets,
comparing classifier performance in two evaluation settings.

We divide the six datasets described in Table~\ref{tbl:existing_datasets}
into training, dev, and test sets. We use the gold-standard
self-report survey data from \citet{preoctiuc2015studying} as our held-out test set
for evaluating all models.
We combine the crowdsourced data from \citet{volkova2015predicting}
and \citet{culotta2015predicting} into a single dataset
containing 3.5k users, which we then split 60\%/40\% to create a training and development set.
The training set is our baseline comparison,
referred to as {\bf Crowd} in our results tables.
We also create class-balanced versions of the dev and test sets with 156 and 452 users,
respectively.
Finally, we use each of our three collected datasets ({\bf QB, HF, CB}) as training sets,
and use a combination of each with the {\bf Crowd} training set.
Thus in total, we have seven training datasets, which make up the bottom seven rows of
our results in Table~\ref{tbl:results}, below.
These results show our three models
evaluated on the imbalanced and balanced test sets.

\begin{table}[!ht]
\small
\centering
\begin{tabular}{c c c c c }
\toprule
     & W    & B    & H/L  & A    \\
\midrule
White           & 12.7 & 4.0  & 3.6  & 4.9  \\
Black           & 3.3  & 16.9 & 1.8  & 3.1  \\
Hispanic/Latinx & 7.6  & 4.0  & 6.5  & 6.7  \\
Asian           & 6.2  & 2.2  & 1.8  & 14.7 \\
\bottomrule
\end{tabular}
\caption{Balanced confusion matrix for BERT on Crowd+{\tt CB}.
Rows show true labels and columns predictions. Each cell
shows test set percentage.}
\label{tbl:bal_confusion}
\end{table}

The balanced and imbalanced dev sets are used for all model and training set combinations in Table~\ref{tbl:results},
which controls for the effect of model hyper-parameter selection.
Cross-validation could be used in practical low-resource settings, but we use a single held-out dev set,
which we subsample in the balanced case.

\subsection{Demographic Prediction Models}
We consider three demographic inference models which we train on each training set.
The first follows \citet{wood2018johns} and uses a single tweet per user.
A character-level CNN maps the user's name to an embedding which is combined with
features from the profile metadata, such as user verification and follower
count. These are passed through a two fully-connected layers to produce classifications.
This model is referred to as ``Names'' in Table~\ref{tbl:results}.
The second model from \citet{volkova2015predicting}
uses a bag-of-words representation of the words in the user's recent tweets as
the input to a sparse logistic regression classifier.
The vocabulary is the 77k non-stopwords that occur at least twice in the dev set.
We download up to the 200 most recent tweets for each user from the Twitter API.
This model is referred to as ``Unigrams'' in Table~\ref{tbl:results}.
The third model uses DistilBERT~\cite{sanh2019distilbert} to embed those same 200 tweets into a
fixed-length representation, which is then passed through logistic regression with L2 regularization to produce a classification.
This model is referred to as ``BERT'' in Table~\ref{tbl:results}.
For all models we tune hyperparameters using the crowdsourced dev set.
Training details for all models are in Appendix~\ref{app:hyperparameter}
and released in our code.

\begin{table*}[!ht]
\setlength{\tabcolsep}{3pt}
\small
\centering
\begin{tabular}{ c c c c c c c c  c }
\toprule
      & {\# Users} & {LD} & {\makecell{CPT}}
      & {TTR}   & {HPT} & {Formality} & {Politeness}  & {Top SAGE Keywords}\\
\midrule
%      users  lexical  contr    TTR        hashtags          formal       polite
A   &  9442 & .751 & .075   & .533   & $.155^{*}$       & -1.770    & .4595 &
liked, visit, hahaha, art, youtube \\
B   & 70838 & .747 & .067   & .532   & $.096^{\dagger}$ & -1.750    & .4584 &
avrillavigne, ni**as, black, ni**a, wit\\
H/L &  8349 & .731 & .051   & .563   & $.145^{*}$       & -1.802    & .4609 &
justinbieber, justin, online, follow\\
W   & 57724 & .759 & .085   & .510   & $.081^{\dagger}$ & -1.697    & .4614 &
bc, realdonaldtrump, snapchat, dog, holy \\
\bottomrule
\end{tabular}
\caption{Comparison of the mean values for each numerical feature between groups.
The last column has the top keywords per group as differentiated according to the SAGE model.
Methods are described in \S~\ref{sec:groups}.  Abbreviations: LD, Lexical
Diversity; CPT, Contractions/tweet; TTR, Type-Token Ratio; HPT, Hashtags/tweet.
Almost all differences are significant; only those numbers that share
superscript symbols are {\bf not} significantly
different at a $0.05$ confidence level when using a Mann-Whitney U test.}
\label{tab:numerical_and_sage}
\end{table*}

\subsection{Evaluation and Baselines}
We consider multiple evaluation setups 
to explore the extreme class imbalance of the survey and crowdsourced datasets
(Table~\ref{tbl:existing_datasets}).
First, we evaluate both total accuracy and macro-averaged F1 score,
which penalizes poor performance on less-frequent classes.
Second, we separately evaluate tuning and testing our models on either imbalanced
or balanced dev and test sets, to see how it affects per-class classifier accuracy.
Finally, we train our unigram and BERT models to reweigh examples with the inverse
probability of the class label in the training data.

We also show the performance of two na\"ive strategies:
randomly guessing across the four demographic categories, and deterministically guessing the majority category.
These baselines highlight the trade-offs between accuracy and F1.
Because the imbalanced test set is so imbalanced, the ``Majority'' baseline strategy can achieve high overall
accuracy, but very low F1. The Random baseline has low overall accuracy but slightly better F1 than the Majority strategy.
These two baselines provide the first two rows of
Table~\ref{tbl:results}.

We stress these evaluation details because the class-imbalance may have
serious implications for downstream applications.
Models trained to do well on the majority class at the expense of minority 
classes could bias downstream analyses by under-representing minority groups.
In public health applications with disparities between groups \cite{laveist2005minority},
not accounting for imbalances between the training and test datasets 
could exacerbate rather than ameliorate inequalities.

\section{Experimental Results and Discussion}

Table~\ref{tbl:results} shows several trends.
The BERT and Unigram models, using 200 tweets per user, generally outperform the single-tweet Names models.
In the imbalanced evaluations, we see a large trade-off between accuracy and F1, with models
achieving higher overall accuracy when they ignore the smaller Asian and Hispanic/Latinx classes.
Even the trivial ``Majority'' baseline is competitive due to the extreme class-imbalance.
While models trained only on {\bf Crowd} achieve significantly higher accuracy
on the imbalanced test set than models trained on our datasets, this is
only because of their excellent performance on White users.
Table~\ref{tbl:class_specific_accuracy} shows the
class-specific accuracy of Unigram models; the model trained only
on the imbalanced {\bf Crowd} dataset achives 95.1\% accuracy on White users, but lower than 50\%, 1\%, and 20\% accuracy on Black, Hispanic/Latinx, and Asian users.
While more
sophisticated approaches to addressing the extreme class imbalance could close
the gap between training on {\bf Crowd} alone and using our noisy datasets, we can
see the benefits of our data in the balanced evaluation. 

\begin{table*}[!ht]
\small
\centering
\begin{tabular}{ l c c c c   c }
\toprule
                                       & Asian    & Black    & Hispanic/Latinx    & White   & Random \\
\midrule
\% users in dataset                    & $6.71$   & $49.44$  & $5.83$   & $38.02$  & -- \\
\midrule
\% users with $1+$ tweets from Android & $38.95^{*\dagger}$  & $38.33^{*}$  & $39.41^{\dagger}$  & $25.46$ & -- \\
\% users with $1+$ tweets from iPhone  & $60.28$  & $58.21$  & $54.89$  & $75.37$ & -- \\
\% users with $1+$ tweets from Desktop & $43.34$  & $30.59$  & $44.87$  & $31.04$  & -- \\
\midrule
\% users with profile URL              & $34.09^{*}$  & $29.71$  & $34.75^{*}$  & $24.78$ & $20.8$ \\
\% users with custom profile image     & $98.83$  & $99.29^{*\dagger}$  & $99.24^{*\ddagger}$  & $99.33^{\dagger\ddagger}$  & $95.4$ \\
\% users with geotagging enabled       & $48.65^{*}$  & $53.27$  & $49.54^{*}$  & $ 56.04$   & $33.1$ \\
\% users with $1+$ geotagged tweet     & $8.35^{*}$   & $6.46$   & $7.81^{*}$   & $ 5.43 $   & $7.9$ \\
\midrule
Average statuses count                 & $11974$ & $18709$ & $12449$ & $14177$ & -- \\
Average tweets per month               & $177.83$   & $255.41$   & $182.13$   & $200.85$ & $739$ \\
\midrule
(m) \% tweets that mention a user      & $59.73$  & $58.71$  & $60.44^{*}$  & $61.77^{*}$ & $22.3$ \\
(m) \% tweets that include an image    & $20.44^{*}$  & $17.20$  & $18.39$  & $19.17^{*}$ & $33.9$ \\
(m) \% tweets that include a URL       & $20.99$  & $21.64$  & $24.01$  & $17.22$  & -- \\
\bottomrule
\end{tabular}
\caption{Profile Behavioral Features.
The first four columns show our {\bf HF} users,
the fifth shows a random sample of 1M users reported in \cite{wood2017does},
when available.
(m) indicates micro-averaging; all others are macro-averaged across users.
Almost all differences between {\bf HF} groups are statistically significant according to a Mann-Whitney U Test.
However, if two entries in the same row share a superscript, they are not significantly different at a $0.05$ confidence level.
We cannot test significance against the random sample.
}
\label{tbl:behavior}
\end{table*}

Across all balanced evaluations, all but one of the models trained with our collected datasets outperform
models trained only on {\bf Crowd} in both accuracy and F1.
Several models improve by more than .10 F1 over models trained only on {\bf Crowd}.
The BERT models achieve the best performance in the balanced evaluation,
while performing relatively poorly on imbalanced data.
This occurs because the BERT models achieve high accuracy on the Black and Asian classes,
which are underrepresented in our imbalanced test set.
We show a confusion matrix for our best balanced model in Table~\ref{tbl:bal_confusion}.

These models are quite simple, and more complex models could improve performance
independent of the dataset.
However, by limiting ourselves to simpler models, we can demonstrate that for
learning a classifier that performs well on four-class classification of race and ethnicity,
our noisy datasets are clearly beneficial.
While the self-reports are noisy, we collect enough data to
support better classifiers on held-out, gold-standard labels.
Despite this experimental improvement, real-world applications
may require more accurate classifiers
or may need to prioritize classifiers with high precision or recall for a particular group.
Such research requires a careful contextualization of what conclusions can be drawn from the available data
and models; classifier error may exaggerate differences between groups.

\section{Twitter Behaviors across Groups} \label{sec:groups}

Our experiments show that our datasets enable better predictive models,
but say nothing about {\em how} self-reporting users use Twitter.
Do different groups in our dataset differ in other behaviors?
We explore this using a variety of quantitative analyses of
Twitter user behavior, following similarly-motivated public health research
~\cite{coppersmith2014quantifying,homan2014toward,gkotsis2016language}.
Two interpretations are possible for these group-level differences:
either user behavior correlates with demographic categories~\cite{wood2017does},
or the {\it choice to self-report} correlates with these behaviors.
These can both be true, and our current methods cannot distinguish between them.
While our empirical evaluation shows that our data is still useful for training
classifiers to predict gold-standard labels, possible selection bias
may influence real-world applications.

Lexical features are widely used
to study Twitter~\cite{pennacchiotti2011machine,blodgett2016demographic}.
For each user in our dataset, we follow \S 3.1 of \citet{inuwa2018lexical} and
calculate Type-Token Ratio\footnote{The number of unique tokens in a tweet
divided by the total number of tokens in the tweet.}, Lexical
Diversity\footnote{The total number of tokens in a tweet without URLs, user
mentions and stopwords divided by the total number of tokens in the
tweet.}~\cite{tweedie1998variable}, and the number of
hashtags and English contractions they use per tweet.
We then use existing trained models for analyzing formality and politeness
~\cite{pavlick2016empirical,danescu2013computational} of online text.
The formality score is estimated with a regression model over lexical and
syntactic features including n-grams, dependency parse, and word embeddings.
The politeness classifier uses unigram features
and lexicons for gratitude and sentiment.
We use the published implementations.\footnote{
\tiny\url{https://github.com/YahooArchive/formality-classifier}}$^\text{,}$\footnote{\tiny\url{https://github.com/sudhof/politeness}}
For both trained models, we macro-average over users' scores to obtain 
a value for each demographic group.
We also use a SAGE~\cite{eisenstein2011sparse} lexical variation implementation to 
find the words that most distinguish each group.
The means of the six quantitative features and the top five SAGE keywords
for each group is shown in Table~\ref{tab:numerical_and_sage}.

We then consider a few basic measures of Twitter usage,
computed from the profile information of each user.
Table~\ref{tbl:behavior} contains the mean value of these features,
describing the broad range of basic user behaviors on the Twitter platform.
Almost all differences in these behavioral features are significant across groups.
Device usage shows the biggest difference; White users are much more likely to have
used an iPhone than an Android to tweet.
In past work, \citet{pavalanathan2015confounds}
demonstrated that the use of Twitter geotagging was more prevalent in metropolitan areas and among
younger users. Table~\ref{tbl:behavior} follows \citet{wood2017does} which
calculated these features for a sample of 1M Twitter users.
Users in our datasets comparatively more often customize their profile
image or URL or enable geotagging. More bots or spam in the random sample 
may partially account for these differences~\cite{morstatter2013sample}.
Table~\ref{tbl:kendall} in Appendix~\ref{sec:more_groups} also compares lists
of the most common common emojis, emoticons, and part-of-speech tags within
each group.

These analyses show substantial differences between the groups labeled by our
self-report methods, suggesting our noisy self-reports 
correlate with actual Twitter usage behavior.  However, it
cannot reveal whether these differences primarily correlate with
racial/ethnic groups or whether these differences appear
from how users decide whether to self-report a race/ethnicity keyword.
Researchers working on downstream public health applications (e.g. \citet{gkotsis2016language})
may want to account for these empirical differences between groups in our
training datasets when drawing conclusions about users in other datasets. 

\section{Limitations and Future Work}
We have presented a reproducible method for automatically identifying self-reports
of race and ethnicity to construct an annotated dataset for training demographic inference models.
While our automated annotations are imperfect,
we show that our data can replace or supplement manually-annotated data.
Our data collection methodology does not rely on large-scale crowd-sourcing,
making it more reproducible and easier to keep datasets up-to-date.
These contributions enable the development and distribution of tools
to facilitate demographic contextualization in computational social science research.

There are several important extensions to consider.
First, our analysis focuses on the United States and English-language racial keywords;
most countries have a unique cultural conceptualizations of race/ethnicity and 
unique demographic composition, and may require a country-specific focus.
We only cover four categories of race/ethnicity,
ignoring smaller populations and multi-racial categories~\cite{jones2001two}.
We use a limited set of query terms, which ignores the diversity of how people may
choose to self-report their identities. While our methods scale easily to additional categories
and/or racial keywords, our evaluation method requires a gold-standard test set
that covers those groups.  For specific applications, a domain expert might
prioritize precision or recall for a specific demographic class. This may involve
fine-tuning a classifier on a dataset constructed with a particular class-imbalance;
the details of that imbalance should be contextualized with the general class
distribution of the population on Twitter.
Our analyses could be compared against human perceptions of users' racial
identity, though past work has suggested such perceptions have underlying
biases~\cite{preoctiuc2017controlling}.  Finally, past work has highlighted
various biases in demographic
inference~\cite{pavalanathan2015confounds,wood2017does},
and our analyses cannot fully rule out the presence of such biases in our data or models.
In future work, we strongly encourage the study of racial self-identities and social cultural issues
as supported by computational analyses.
These issues should be viewed from a global perspective,
especially with regards to biases in our collection methods~\cite{landeiro2016robust}.

We release our code in the {\tt
Demographer} package to enable training new models and constructing future
updated datasets. We also release our trained models and annotated Twitter
user ids for academic researchers that agree to the data use
agreement and obtain approval from an ethics board.

\clearpage

\bibliography{paper}
\bibliographystyle{acl_natbib}

\clearpage

\appendix

\section{Preprocessing, Tokenizing, and Tagging} \label{sec:preprocessing}
We lowercase all descriptions and use NLTK Tweet Tokenizer \cite{bird2009natural} to get the PoS tags.
Our candidate self-report words are scraped from 177M Twitter descriptions using the regex and PoS pattern,
\texttt{\{I`/I a\}m (+ RB)( + DT) (+ JJ) + NN}.
We collect both adjectives and nouns from the pattern above, and refine the matches
by keeping adjectives and nouns that match the majority tag in the Google N-gram corpus.
We filter out plural words (e.g. ``white {\bf people}'') using a PoS tag pattern,
\texttt{JJ + NNPS/NNS}, and refer to our set of self-report words as $S$.

\section{Calculating the ``Self-Report'' Score}\label{sec:self-report}
To calculate the score described in \S~\ref{sec:data_collection},
we first obtain simple co-occurrence weighting by counting
the occurrences $O_s(w_s)$ of word $w_s$ as a self-report word, and
its overal occurrences $O(w_s)$. Then:
\[
R = \sum_{w_s \in S^{win}}\frac{1}{D(w_s,q)} \cdot \frac{O_s(w_s)}{O(w_s)},
\]
where $S^{win}$ is the self-report words in the fixed window size,
$D(w_s,q)$ denotes the distance between $w_s$ and query word $q$.

We also consider a TF-IDF weighting as:
\begin{align*}
R_{\text{tfidf}} = \sum_{w_s \in S^{win}}\frac{1}{D(w_s,q)} & \cdot \frac{O_s(w_s)}{O(w_s)} \\
  &\cdot \log \frac{\sum_{w \in S}O_s(w)}{O_s(w_s)}
\end{align*}

To fine-tune our self-report score, three authors manually labeled a tuning set
of 400 descriptions as to whether the user was self-reporting a matching query word,
using a three-label nominal scale of ``yes,'' ``no,'' and ``unsure.'
We discarded 6 that we classified as organizations~\cite{wood2018johns},
and had an Krippendorff $\alpha$ $0.8058$ on the remaining 394.
We use majority voting strategy to get binary labels and
select the self-report score's hyperparameters of window size and threshold,
and whether to use the tf-idf weighting, based on the precision calculated on this tuning set.

To ensure that these chosen hyperparameters did not overfit to the tuning set,
we sampled an additional $199$ users from {\bf HF}.
Using a three-label nominal scale of ``yes,'' ``no,'' or ``unsure,'' the three
annotators achieved a Krippendorff's alpha of $0.625$. 
After converting to binary ``yes'' and ``no'' by taking majority voting
and discarding $7$ users who were majority ``unsure,''
our best model achieves $72.4\%$ accuracy on the test set
with simple weighting, window size 5, and threshold of 0.35.

\section{Model Training Details} \label{app:hyperparameter}

Our name model uses a CNN implementation released in \citet{wood2018johns}.
We use a CNN with 256 filters of width 3.
The user's name (not screen name) is truncated at 50 characters
and embedded into a 256 dimensional character embedding.
We fine-tuned the learning rate on our dev data,
trained for 250 epochs, and used early-stopping
on dev-set F1 to pick which model to evaluate on the test set.

Our unigram model follows \citet{volkova2015predicting},
using a simple sparse logistic regression.
We use an implementation from Scikit-Learn, and tune the
regularization parameter on the dev set.
We introduce a hyperparameter to down-weight the contribution
of our users compared to the baseline users; 
we also set that parameter on the dev set.

For BERT model, we first get embedding for every tweet
by taking the vector with size 768 on special [CLS] token in the last hidden layer. 
The element-wise average of all tweet embeddings from one user is then passed through 
a logistic regression model with L2 regularization to make the classification.
Similarly, the regularization parameter is tuned on the dev set.
We fine-tuned DistilBERT model on tweets collected from training set split of the crowdsourced dataset.
However, after observing limited performance improvement we just use pre-trained DistilBERT model.

\begin{table}[!ht]
\setlength{\tabcolsep}{3pt}
\small
\centering
\begin{tabular}{ l  c c c c c c c c c c c l }
\toprule
            & \multicolumn{2}{c }{Emojis}  & \multicolumn{2}{c }{Hashtags} & \multicolumn{2}{c }{PoS bigrams}  \\
Top $k$     & {20}  & {50} & {20}  & {50}   & {20}  & {50}   \\
\midrule
A v. B   & -0.67 & -0.26   & -0.85 & -0.87  & 0.29  & 0.19   \\
A v. H/L & -0.10 & -0.07   & -0.84 & -0.86  & 0.55  & 0.02   \\
A v. W   & -0.38 & 0.13    & -0.83 & -0.80  & 0.02  & -0.02  \\
B v. H/L & -0.65 & -0.38   & -0.83 & -0.82  & 0.52  & 0.03   \\
B v. W   & -0.48 & -0.16   & -0.79 & -0.72  & 0.04  & 0.24   \\
H/L v. W & -0.40 & -0.13   & -0.91 & -0.89  & -0.17 & -0.28 \\
\bottomrule
\end{tabular}
\caption{Kendall's $\tau$ correlation coefficients for top items of different list features.
For hashtags in particular we see large negative coefficients.}
\label{tbl:kendall}
\end{table}

\section{Additional Analyses of Twitter Behavior across Groups} \label{sec:more_groups}

This appendix contains an additional analysis following \S~\ref{sec:groups}.

In addition to the SAGE keyword comparison, we 
explore topical differences between groups by compiling ranked lists
of common emojis, emoticons, and part-of-speech tags within each group.
Table~\ref{tbl:kendall} shows a comparison of Kendall $\tau$ rank correlation
between these To compare across groups, we look at the top $k$ items in each
list and calculate Kendall $\tau$ rank correlation coefficients for each pair
of demographic groups~\cite{morstatter2013sample}.  Table~\ref{tbl:kendall}
shows pairwise $\tau$ correlations. These coefficients vary
between -1 and 1 for perfect negative and positive correlations.  For emojis,
all correlations are negative for $k=20$, but increase at $k=50$.  For
hashtags, however, correlations are strongly negative for all values of $k$,
suggesting that groups labeled by our method substantially differ in the topics
they discuss.  While we use English keywords for data collection, topic
difference may be confounded by users' native language(s).

\section{Data Statement}

Following \citet{bender2018data}, we highlight characteristics of our collected noisy self-report
data that may be important for mitigating ethical and scientific missteps.

\paragraph{Curation rationale} Examples of Twitter users who self-report their racial identity
using English-language keywords.

\paragraph{Language variety} While our dataset contains predominantly English (en-US),
there is substantial diversity in language due to the international and 
due to the informal setting of Twitter.
When we randomly sample 1000 users from our {\tt Heuristic Filter} list and consider up to 100 tweets
per user, we find that the Twitter-produced {\tt lang} field indicates that 78.5\% of the tweets
are in English, with the next three most-common {\tt lang} labels as Spanish (3.8\%),
Portuguese (3.7\%), and Undetermined (3.3\%).

\paragraph{Speaker demographics} The speakers in our dataset are Twitter users.
To be included in our initial dataset, users must use an English racial self-report keyword
in their Twitter profile description, and must not be labeled as an organization
by the classifier from \citet{wood2018johns}.
We then perform additional filtering of users, detailed in the paper, to improve the likelihood
that a racial self-report keyword is actually self-reporting race.

\paragraph{Annotator demographics} Our small manual annotation was conducted by
three authors, Asian and White men, ages 20-30, with native languages of Chinese and English.

\paragraph{Speech situation} Twitter user profiles and tweets.

\paragraph{Text characteristics} Informal Twitter user descriptions and tweets. We
make no restrictions on the content of the tweets.

\end{document}